%% file: main.tex
\documentclass[conference]{IEEEtran}
\IEEEoverridecommandlockouts
\usepackage{comment}
\usepackage{amsmath,amsfonts}
\usepackage{array}
\usepackage[caption=false,font=normalsize,labelfont=sf,textfont=sf]{subfig}
\usepackage{textcomp}
\usepackage{stfloats}
\usepackage{url}
\usepackage{verbatim}
\usepackage{graphicx}
\usepackage{array}
\usepackage{caption} 
\usepackage{xcolor}
\usepackage{multirow}
\usepackage[ruled, lined, linesnumbered, longend]{algorithm2e}
\usepackage{soul} 

\def\BibTeX{{\rm B\kern-.05em{\sc i\kern-.025em b}\kern-.08em
    T\kern-.1667em\lower.7ex\hbox{E}\kern-.125emX}}
    
\newcommand{\fixme}[1]{{\color{red}\em\bf{[FIXME: #1]}}}

\SetKwInput{KwOutput}{Output}   
\SetKwInput{KwInput}{Input}

\newcommand{\shakshi}[1]{{\color{black} {\textbf{}}\color{black}{#1}}{\color{black}}}

\begin{document}

\title{AMIR: An Automated MisInformation Rebuttal System --- A COVID-19 Vaccination Datasets based Exposition}


\author{
\IEEEauthorblockN{Shakshi Sharma}
\IEEEauthorblockA{\textit{School of Artificial Intelligence} \\
Bennett University, Greater Noida, India\\
shakshi.sharma268@gmail.com}
\and
\IEEEauthorblockN{Anwitaman Datta}
\IEEEauthorblockA{\textit{College of Computing and Data Science} \\
Nanyang Technological University, Singapore\\
anwitaman@ntu.edu.sg}
\and
\IEEEauthorblockN{Rajesh Sharma}
\IEEEauthorblockA{\textit{Institute of Computer Science} \\
University of Tartu, Estonia\\
rajesh.sharma@ut.ee}
}

\maketitle

\begin{abstract}
Misinformation has emerged as a major societal threat in the recent years in general; specifically in the context of the COVID-19 pandemic, it has wrecked havoc, for instance, by fuelling vaccine hesitancy. Cost-effective, scalable solutions for combating misinformation are the need of the hour. This work explored how existing information obtained from social media and augmented with more curated fact checked data repositories can be harnessed to facilitate automated rebuttal of misinformation at scale. While the ideas herein can be generalized and reapplied in the broader context of misinformation mitigation using a multitude of information sources and catering to the spectrum of social media platforms, this work serves as a proof of concept, and as such, it is confined in its scope to only rebuttal of tweets, and in the specific context of misinformation regarding COVID-19. It leverages two publicly available datasets, viz. FaCov (fact-checked articles) \cite{sharma2022facov} and misleading (social media Twitter) \cite{sharma2022mis} data on COVID-19 Vaccination\footnote{This work has been accepted at IEEE Transactions on Computational Social Systems. Please cite the official version}.

\end{abstract}




\input{Introduction}

\input{related}


\input{Approaches_datasets}

\input{Proactive_approach}

\input{Factcheck_approach}

\input{conclusions}

\section*{Acknowledgments}
R.Sharma's work has received funding from the EU H2020 program under the SoBigData++ project (grant agreement No. 871042), by the CHIST-ERA grant CHIST-ERA-19-XAI-010,  ETAg (grant No. SLTAT21096), and partially funded by HOLISTIC ANALYSIS OF ORGANISED MISINFORMATION ACTIVITY IN SOCIAL NETWORKS project (PCI2022-135026-2).


\bibliographystyle{IEEEtran}
\bibliography{main}

\end{document}

%% file: Introduction.tex
\section{Introduction}\label{sec:intro}

The menace of misinformation, with accelerated dissemination through social media and amplified in echo chambers, poses grave risks. A mix of technical, social and regulatory policies together is needed to thwart these. In that context, technical mechanisms to rebut misinformation in a timely manner at scale, customized to the specifics of the misinformation and concerned users have a crucial role to play. To that end, we consider automated misinformation rebuttal systems are essential.

In this work, we explore two complementary approaches to carry out rebuttal of misinformation in the context of the Twitter platform. The first approach repurposes existing social media content, in which we identify and reuse existing related but factual/non-misleading tweets by other users as recommended counter tweets. This approach can be used even when authorative entities and fact-checking websites are yet to carry out any vetting exercise for a given misinformation, or while such articles are yet to be ingested and matched by our second approach.

The second approach leverages fact-checked articles available from several popular fact-checking websites. This approach aims to provide users with verified, reliable, and more comprehensive information from reputable sources to counter the misinformation. By utilizing both of these automation approaches in tandem, we create a scalable rebuttal pipeline to combat misinformation. \shakshi{We note that the currently proposed approach requires some extent of human intervention during the initial model building phase, and similar intermittent interventions may also be necessary to update the models in the background. As such, the automation and its benefits are during the operational phase of the system, i.e., the rebuttals recommendation is automated.} 

While there are numerous fact-checking websites, to the best of our knowledge, such augmentation of fact-checked articles matched to individual misinformation content, to be recommended for automated rebuttal, is unique, and closes a vital gap, since individual users may not even realize that they need to fact-check a given piece of (mis-)information they are exposed to, and even if they did want to do so, they might not be able to look for and locate appropriate information from credible sources.

The purpose of this work was to demonstrate the feasibility and practicality of a reasonably simple and thus easy to implement, deploy, generalize approach. As such, while our methodology is adequately general, and can be applied to a wider variety of misinformation or even a more diverse range of social media platform content, this work is confined to the use of COVID-19 specific datasets, in part because it is a timely topic in need of immediate attention, and in part because of the availability of relevant curated data in abundance. Specifically, we use \cite{sharma2022facov} as the corpus of fact-checked articles, and \cite{sharma2022mis} as the corpus of tweets with labels for being misinformation or not.

The rest of the paper is organized as follows. Section \ref{sec:related} reviews the literature.
Section \ref{sec:define_approaches_datasets} discusses the two approaches to combat misinformation in brief along with datasets and the preparatory steps used for evaluating the recommendation approaches. Section \ref{sec:proactive_recommendations} details how we repurpose existing relevant non-misleading social media content for rebuttal. Section \ref{sec:approach} explains the steps used to identify and recommend relevant fact-checked article(s).
Finally, Section \ref{sec:conclusion} concludes with the implications of the presented work and our future plans.

%% file: related.tex
\section{Related Work}\label{sec:related}

Previous studies \cite{balakrishnan2022infodemic} have identified various factors that contribute to the spread of false news, such as low factual understanding and difficulty recognizing fake news. To combat misinformation, providing correct information to users is crucial. While many studies have focused on detecting misinformation and interpreting black-box models \cite{butt2022goes,sharma2022mis},
combating fake news in an automated manner has received relatively less attention so far \cite{dhawan2024game}.

\subsection{Combating Misinformation}

There are various network-based approaches for mitigating the spread of fake news on social media \cite{kempe2003maximizing,nguyen2012containment,sharma2019combating,zhou2013learning,farajtabar2017fake,sharma2021identifying}. However, these methods have limitations such as a lack of external moderation and a reliance on strict assumptions that fake news has already been identified and its propagation is tracked. Additionally, these methods are difficult to deploy on real-world social networks due to the dynamic and volatile nature of information diffusion and user behavior. 

Deep learning based approaches utilize advanced neural network architectures, such as recurrent neural networks (RNNs) and convolutional neural networks (CNNs), to model the text and context of news articles \cite{shahid2022detecting,sharma2021identifying}, and exploring more advance techniques such as knowledge graphs \cite{mayank2022deap} and reinforcement learning \cite{nikopensius2023reinforcement}. These models can be trained on large datasets of labeled news articles to learn the underlying patterns and features that distinguish fake news from real news. 
Once trained, these models can be used to classify new articles as fake or real with varying degrees of accuracy. Additionally, these models can also be used to extract features from news articles that can be used as inputs for other mitigation strategies, such as URL or news recommendation \cite{sharma2023misinformation}. However, these approaches also have limitations, such as the need for large amounts of labeled training data and are not robust against biases in the training data.

There have been some efforts to use recommender systems to combat fake news, primarily focused on recommending fact-checked URLs to a small specific group of users, e.g., fact checkers \cite{vo2018rise,you2019attributed}. 

\subsection{COVID-19 Misinformation \& Mitigation}
Our study uses COVID-19 as a case, so we present literature on COVID-19 misinformation.
The studies \cite{skafle2022misinformation,lieneck2022facilitators} show that misinformation about COVID-19 vaccines on social media platforms can lead to hesitancy and a decrease in vaccine uptake. 
There are works broadly focused on three aspects: techniques for detecting fake news, characterizing the misinformation circulating, and strategies for combating misinformation. 
Techniques for detecting fake news include text-based, image-based, and network-based methods, using machine learning and deep learning algorithms to identify patterns that are characteristic of misinformation \cite{almalki2020health,mayank2022deap,jagtap2021misinformation,monti2019fake,elhadad2019fake}. 
Characterizing misinformation includes identifying common themes, sources, and platforms for misinformation \cite{sharma2022facov,ngai2022impact,puri2020social}. 
Strategies for combating misinformation include fact-checking, education campaigns, and interventions on social media platforms \cite{sallam2021covid}. 


Other strategies include tracking vaccine misinformation in real-time and engaging with social media to disseminate correct information \cite{islam2021covid}, semantic network analysis \cite{melton2021network} which revealed that the most common topics of vaccine misinformation were related to safety, efficacy, and conspiracy theories, and that misinformation was spread by a small number of users who were highly active in the anti-vaccination movement. 


The work that is closest to ours \cite{WangclosestWWW2022} focuses on utilizing the user's reading history to identify the event or topic they are interested in. However, it may struggle to identify new or breaking events that are not yet a part of the user's reading history. This means that the approach may not be able to provide the user with timely and relevant information about new or unexpected events. 
Another limitation of the approach is that it relies solely on fact-checked articles for recommendations. While fact-checked articles are a reliable source of information, they are not always immediately available. Misinformation can spread quickly, often before fact-checking organizations have a chance to verify the information and publish their findings. This means that the approach may not be able to provide recommendations for news that is being widely shared and discussed in real-time, which is when users are most vulnerable and also likely to be searching for information about it.

In our work, we have developed a pipeline to automate the rebuttal of misinformation on social media. The pipeline combines two approaches, the first approach is identifying and reusing relevant existing but factual/non-misinformation tweets as counter tweets. This is the baseline approach which can be used even when fact-checking websites are yet to carry out the vetting exercise for a given misinformation, or while such articles are identified and matched by our second approach. 
The second approach is recommending fact-checking articles from various fact-checking websites when they are available. This pipeline creates a more practical and scalable rebuttal system to combat misinformation on social media, by providing users with verified, reliable and more comprehensive information from reputable sources.



%% file: Approaches_datasets.tex
\section{Combating Misinformation: A Two-Pronged Approach}\label{sec:define_approaches_datasets}

\begin{figure*}[!htbp]
    \centering
    \includegraphics[width=14cm, height = 5cm, trim = {0, 5cm, 0cm, 1.5cm}]{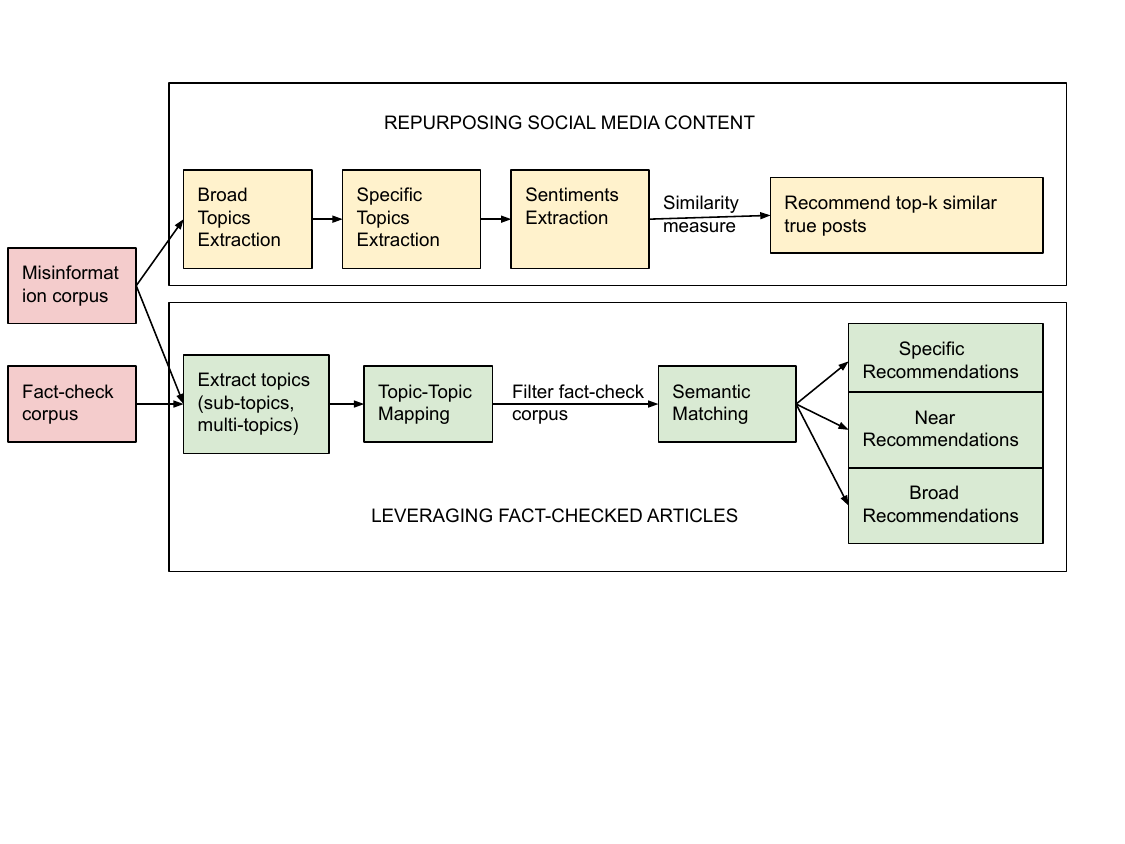}
    \caption{Automated Misinformation Rebuttal Pipeline.}
    \label{fig:proposed}
\end{figure*}

In order to automate the rebuttal of misinformation on social media, we have developed a pipeline that utilizes two complementary approaches (as shown in Figure \ref{fig:proposed}). The first identifies and reuses existing related but factual/non-misleading tweets by other users as recommended counter tweets. This is the baseline approach which can be used even when fact-checking websites are yet to carry out the vetting exercise for a given misinformation, or while such articles are identified and matched by our second approach. 

The second recommends fact-checking articles (subject to availability) from various fact-checking sites. This provides users with verified, reliable and more comprehensive information from reputable sources to counter the misinformation. By utilizing both of these automation approaches in tandem, we create a scalable rebuttal pipeline to combat misinformation on social media.

Note that in a live deployment of our approach, a module to carry out the classification of tweets into misinformation (or not) categories would be needed, for which various options exist, e.g., \cite{sharma2022mis,li2022dynamic,bhowmik2022novel,koloski2022knowledge,mosallanezhad2022domain,chen2022multi}. Such classification of content is outside the scope of this work, as such, we used the pre-labeled data from \cite{sharma2022mis}.

Though instantiated with COVID-19 specific (mis-)information on Twitter, the applied methodology and principles are universal and generalizable, subject to training the models with suitable datasets. Likewise, the pipeline (see Figure \ref{fig:proposed}) 
is inherently modular in nature, such that, for a given individual step or task, alternative approaches can be applied separately, or even in conjunction.

\subsection{Datasets for AMIR instantiation}

We utilized two publicly available COVID-19 datasets to instantiate the proposed Automated MisInformation Rebuttal (AMIR) System. The first dataset, FaCov \cite{sharma2022facov}, is an extensive corpus of fact-checked articles collected from 13 fact-checking websites over a period of two years, from December 2019 to the first week of December 2021. The second dataset \cite{sharma2022mis} is a collection of tweets discussing the COVID-19 vaccine, extracted from Twitter, which was labeled as misinformation or not. We refer readers to the respective papers for more information on the datasets. 
While both datasets identify key themes surrounding the COVID-19 pandemic, the FaCov dataset also uncovers sub-topics, which we leverage in our analysis.
Next, we delve into the information extracted from the datasets, which serves as the foundation for our recommendation approaches.

\subsection{Data Preparation: Discerning (sub-)topics}

We extend the analysis of the data from \cite{sharma2022facov,sharma2022mis} to examine topics at a finer granularity and study the relationships among them. 

\subsubsection{Social media (Twitter) corpus \cite{sharma2022mis}}\label{sec:social_topics}

We use Latent Dirichlet Allocation (LDA) topic modeling to assign topics to each tweet in the  misleading corpus \cite{sharma2022mis}. This refines the prior work in \cite{sharma2022mis} by assigning specific topic(s) to individual tweets rather than simply identifying broad topics within the dataset. 
This resulted in 12 identified topics, with a small number of tweets remaining unlabeled. To improve the labeling process for these tweets, we also checked for the presence of synonyms for each topic in the unlabeled tweets and were able to assign some tweets to the previously identified topics. Despite this, a small fraction (2.98\%) of tweets remain unlabeled (labeled as \textbf{\textcolor{gray}{Unknown}}) shown in Table \ref{tbl:topics}.

\begin{table}[!htbp]
\footnotesize
\centering
\caption{Absolute number (and \%) of Tweets per topic. 
}
\label{tbl:topics}
\resizebox{0.66\columnwidth}{!}{%
\begin{tabular}{|l|l|}
\hline
\textbf{Topics}      & \textbf{Tweets: number (\%)} \\ \hline
Choices              & 33,150 (28.9\%) \\ \hline
Politics             & 25,276 (22\%)   \\ \hline
Vaccine Efficacy     & 21,936 (19.1\%) \\ \hline
Shots                & 9,568 (8.34\%)  \\ \hline
Trump                & 8,432 (7.35\%)  \\ \hline
Data \& Facts        & 3,601 (3.14\%)  \\ \hline
\textbf{\textcolor{gray}{Unknown}}              & \textcolor{gray}{3,426 (2.98\%)}  \\ \hline
Trials               & 3,217 (2.8\%)   \\ \hline
Myths                & 2,376 (2\%)     \\ \hline
Operation Warp Speed & 1,369 (1.19\%)  \\ \hline
Real Side-Effects    & 1,216 (1.06\%)  \\ \hline
Approval             & 883 (0.77\%)    \\ \hline
Availability         & 185 (0.16\%)    \\ \hline
\end{tabular}%
}
\end{table}

\begin{table*}[!htbp]
\centering
\caption{Topics and Subtopics identified from the Misleading tweets dataset \cite{sharma2022mis}. 
}
\label{tab:misleading_subtopics}
\resizebox{0.8\textwidth}{!}{%
\begin{tabular}{|l|l|l|l|}
\hline
\textbf{Main topic}           & \textbf{Sub-topic 1}                                                                                         & \textbf{Sub-topic 2}                                                                                      & \textbf{Sub-topic 3}                                                                                   \\ \hline
Politics             & Operation Warp Speed                                                                                & Vaccine Efficacy                                                                                 & Vaccine in countries                                                                          \\ \hline
Vaccine Efficacy     & Vaccine Effects on pregnant women                                                                   & Trust \& risk of the vaccines                                                                    & \begin{tabular}[c]{@{}l@{}}vaccine-related data \\ (deaths, illness, shot, dose)\end{tabular} \\ \hline
Vaccine Choices              & MRNA vaccines (allergies)                                                                           & Orders of vaccines                                                                               & Individual’s interest on vaccines                                                             \\ \hline
Vaccine Shots                & Second doses received in time                                                           & Single doses received in time                                                        & -                                                                                             \\ \hline
Operation Warp Speed & \begin{tabular}[c]{@{}l@{}}White house providing funding \\ to manufacturing companies\end{tabular} & \begin{tabular}[c]{@{}l@{}}Money involved in providing \\ doses of various vaccines\end{tabular} & \begin{tabular}[c]{@{}l@{}}Administration, distribution of \\ approved vaccines\end{tabular}  \\ \hline
Real Side-effects    & Fever \& soreness after Pfizer shots                                                        & Soreness and pain after shots                                                            & Experiencing fatigue \& headache                                                              \\ \hline
Trump                & \begin{tabular}[c]{@{}l@{}}Involvement of Trump in\\  the Operation Warp Speed\end{tabular}         & -                                                                                                & -                                                                                             \\ \hline
Data \& Facts        & Report shots of vaccines                                                                            & -                                                                                                & -                                                                                             \\ \hline
Myths                & Bill gates conspiracy theory                                                                        & Deaths \& diseases caused by vaccines                                                            & Severe allergies due to vaccines                                                              \\ \hline
Trials               & Phase trials of various vaccines                                                                    & Placebo effects                                                                                  & Participants facing illnesses                                                                 \\ \hline
Vaccine Approval             & -                                                                                                   & -                                                                                                & -                                                                                             \\ \hline
Vaccine Availability         & -                                                                                                   & -                                                                                                & -                                                                                             \\ \hline
\end{tabular}
}
\end{table*}
We similarly identify up to three sub-topics of the main topics in the misleading corpus (shown in Table \ref{tab:misleading_subtopics}). The number of topics and the sub-topics have been chosen based on the high coherence value in the LDA technique. 
There are a couple of topics, viz. \textbf{Trump} and \textbf{Data \& Facts}, wherein only one sub-topic is identified. Two topics, namely, \textbf{Approval} and \textbf{Availability} have no further sub-topics.

\subsubsection{Fact-checked articles corpus \cite{sharma2022facov}}

Table \ref{tbl:facov_subtopics_mapping_misleading_topics} shows the most discussed topics and corresponding sub-topics in the fact-checked articles dataset \cite{sharma2022facov}, capturing a variety of issues including posts involving Trump, vaccine-related concerns, the number of deaths due to COVID-19, falsely relating images or videos to COVID-19.

It's interesting to note that the topics identified in the misleading Twitter corpus are more fine-grained and nuanced as compared to fact-checked articles FaCov corpus. For instance, on the topic of \textbf{Vaccine Efficacy}, the three sub-topics delve into its effect on pregnant women, concerns about the vaccines, and statistics about the vaccine. Whereas in the case of the FaCov dataset, the only sub-topic of the \textbf{Vaccine's effects} is COVID preventive measures. 
This is likely because Twitter is a platform where users openly express their concerns and perspectives, whereas, in the fact-checked articles FaCov dataset, only specific posts flagged by many people are fact-checked and posted on fact-checking websites. This is in line with observations drawn from other social media platforms such as Reddit \cite{redditzachary}, where it has also been noted that there are more nuances and diversity in user generated content in the form of comments and discussions, with respect to website atricle subjects. 

\begin{table}[!htbp]
\centering
\caption{Topics and subtopics identified from the FaCov dataset \cite{sharma2022facov}. Furthermore, items highlighted in \textcolor{blue}{Blue} color represents the mapped topics from the misleading tweets dataset \cite{sharma2022mis}.
}
\label{tbl:facov_subtopics_mapping_misleading_topics}
\resizebox{\columnwidth}{!}{%
\begin{tabular}{|l|l|l|l|}
\hline
\textbf{Main topic}                                                                                                        & \textbf{Sub 1}                                                                   & \textbf{Sub 2}                                                                       & \textbf{Sub 3}                                                                                 \\ \hline
School e-learning                                                                                                  & \begin{tabular}[c]{@{}l@{}}Republican, Democrat\\ interviews\end{tabular}        & COVID cases                                                                          &                                                      -                                          \\ \hline
\begin{tabular}[c]{@{}l@{}}Trump posts on health\\ Workers \textcolor{blue}{(Trump)}\end{tabular}                                            & \begin{tabular}[c]{@{}l@{}}Trump’s interview on\\ health workers\end{tabular}    & \begin{tabular}[c]{@{}l@{}}Social media posts on\\ COVID cases\end{tabular}          &        -                                                                                        \\ \hline
\begin{tabular}[c]{@{}l@{}}Vaccine’s effects \textcolor{blue}{(Vaccine Efficacy)}\\  \textcolor{blue}{(Vaccine Shots) (Vaccine Choices) (Myths)}\\ \textcolor{blue}{(Vaccine Availability)}\end{tabular} & \begin{tabular}[c]{@{}l@{}}COVID preventive\\ measures\end{tabular}              &   -                                                                                   &                                                -                                                \\ \hline
\begin{tabular}[c]{@{}l@{}}Coronavirus\\ death/cases in states \textcolor{blue}{(Politics)}\end{tabular}                                     & \begin{tabular}[c]{@{}l@{}}Misleading posts,\\ pictures, and videos\end{tabular} &                    -                                                                  &        -                                                                                        \\ \hline
\begin{tabular}[c]{@{}l@{}}Trump and Biden\\ claims on China \textcolor{blue}{(Politics) (Trump)}\end{tabular}                               & \begin{tabular}[c]{@{}l@{}}Democrats discussion\\ on COVID reports\end{tabular}  & \begin{tabular}[c]{@{}l@{}}Trump and Biden\\ elections amid \\ pandemic\end{tabular} &          -                                                                                      \\ \hline
\begin{tabular}[c]{@{}l@{}} Masks \textcolor{blue}{(Myths)} \textcolor{blue}{(Trials)}  \\\textcolor{blue}{(Data \& Facts)}               \\\end{tabular}                                                                                              & \begin{tabular}[c]{@{}l@{}}Misleading posts,\\ pictures, and videos\end{tabular} & \begin{tabular}[c]{@{}l@{}}Drug trials and\\ Immunity \end{tabular}          & \begin{tabular}[c]{@{}l@{}}China's COVID\\ reports and research\\\end{tabular} \\ \hline
\begin{tabular}[c]{@{}l@{}}Spread of COVID-19 \\ \textcolor{blue}{(Politics) (Myths)}\end{tabular}                                           & False death reports                                                              & Masks                                                                                & \begin{tabular}[c]{@{}l@{}}False videos on\\ vaccines\end{tabular}                             \\ \hline
\begin{tabular}[c]{@{}l@{}}Misleading posts,\\ pictures, and videos\\ \textcolor{blue}{(Myths)}\end{tabular}                                 & Lockdown                                                                         &                   -                                                                   &            -                                                                                    \\ \hline
\end{tabular}
}
\end{table}

\subsection{Data Preparation: Relationships Among Topics in Fact-check Article Corpus}

\begin{table}[!htbp]
\centering
\caption{(Multilabel topics) Topmost topics distribution in fact-checked articles FaCov dataset. The Count column indicates the total number of fact-checked articles that contains the specific topic.}
\label{tbl:topic_distribution_1}
\resizebox{0.66\columnwidth}{!}{%
\begin{tabular}{|l|l|}
\hline
\textbf{Topics}                        & \textbf{Count} \\ \hline
Vaccine Effects                        & 967            \\ \hline
Trump posts on health Workers          & 664            \\ \hline
Misleading posts, pictures, and videos & 577            \\ \hline
Coronavirus death/cases in states      & 461            \\ \hline
School e-learning                      & 306            \\ \hline
Spread of COVID-19                     & 81             \\ \hline
Masks                                  & 32             \\ \hline
\end{tabular}
}
\end{table}

\begin{table}[!htbp]
\centering
\caption{(Multilabel topics) Secondmost topics distribution in fact-checked articles FaCov dataset. The Count column indicates the total number of fact-checked articles that contains the specific topic.
}
\label{tbl:topic_distribution_2}
\resizebox{0.66\columnwidth}{!}{%
\begin{tabular}{|l|l|}
\hline
\textbf{Topics}                        & \textbf{Count} \\ \hline
Trump posts on health Workers          & 834            \\ \hline
Misleading posts, pictures, and videos & 635            \\ \hline
Coronavirus death/cases in states      & 558            \\ \hline
Vaccine Effects                        & 458            \\ \hline
School e-learning                      & 315            \\ \hline
Masks                     & 137            \\ \hline
Spread of COVID-19                     & 130            \\ \hline
Unknown                     & 21           \\ \hline
\end{tabular}
}
\end{table}

We investigate co-occurrence of topics for each fact-checked article, considering the second most suitable topic among the same topics using the probability distribution from the LDA approach. 
In Tables \ref{tbl:topic_distribution_1} and \ref{tbl:topic_distribution_2} we show the distribution of the top two topics assigned to each article. The \textit{Count} column indicates the total number of fact-checked articles that contains the specific topic. 
The widely discussed topics are the \textbf{Vaccine Effects} and \textbf{Trump posts on health Workers} in the topmost and second most topics distribution, indicating that the majority of the fact-checked articles are related to clarifying the efficiency and symptoms of the various COVID-19 vaccines. The \textbf{Trump posts on health workers} topic also gained popular traction in false claims.

\begin{figure}[!htbp]
    \centering
    \includegraphics[width = 8cm, height = 5cm]{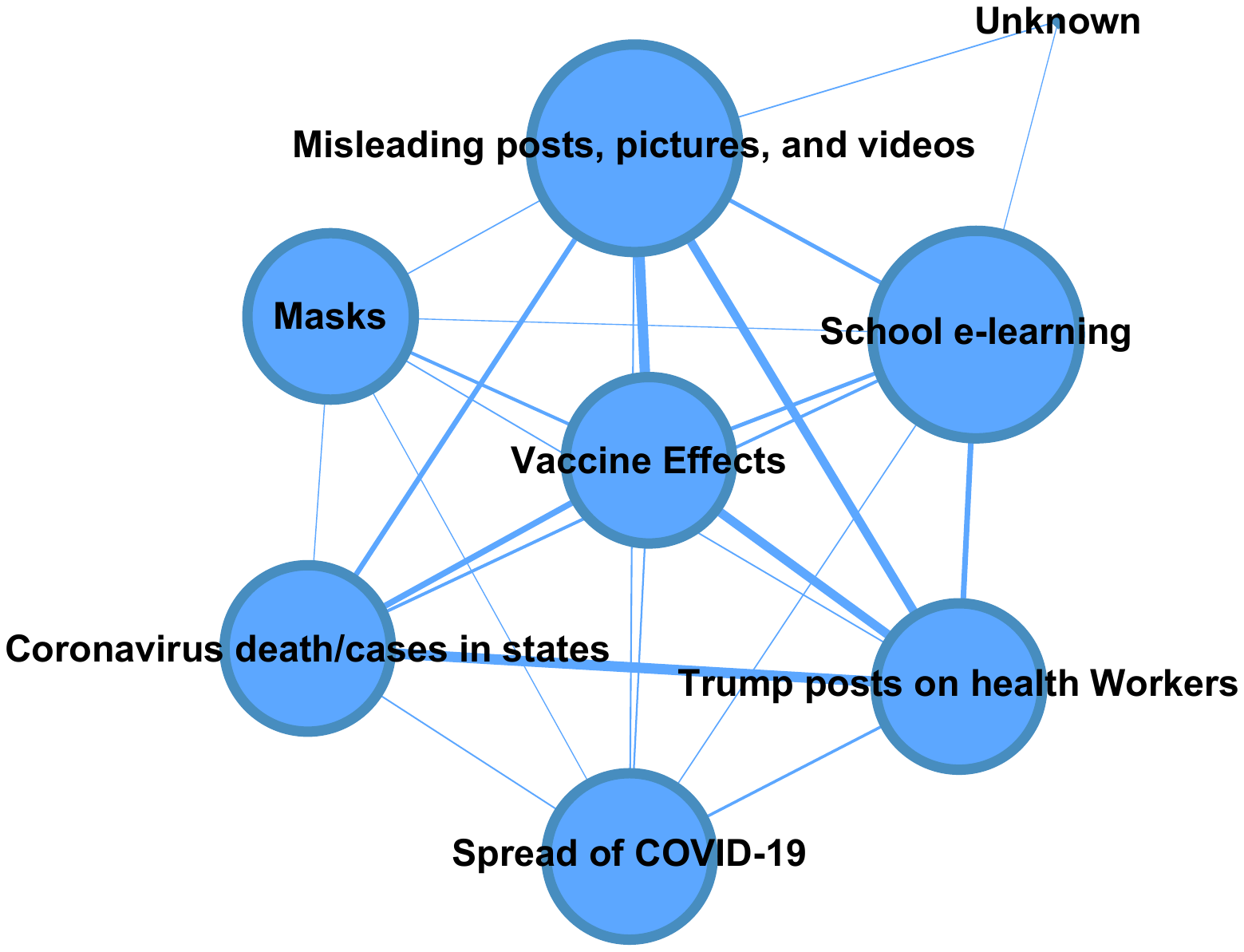}
    \caption{Co-occurrence of prominent topics in the corpus of fact-checking articles. 
    }
    \label{fig:graph_example}
\end{figure}

\shakshi{Figure \ref{fig:graph_example} shows the relations between prominent topics. The nodes and edges of the graph represent the topics' names and whether both the topics have been assigned to the same article (based on the multilabel topics as discussed above), respectively. The size of the node depends upon the degree of the node. The edges' weight (indicated with thickness in the figure) encapsulates the number of times the same two topmost topics are assigned to an article. }

We identify four relatively strong relations, namely, \textbf{Coronavirus death/cases in states --- Trump posts on health workers;  Vaccine Effects ---  Misleading posts, pictures, and videos; Vaccine Effects --- Trump posts on health workers; Trump posts on health workers  ---  Misleading posts, pictures, and videos}.
These show that the fact-checked articles mostly present the number of COVID-19 (death) cases within the topics with Trump's posts on health workers. Vaccine-related concerns has been widely linked with misleading posts, images or videos which are wrongly linked to COVID-19. In addition, Trump's posts on health workers are linked with misleading posts, images or videos.

Including the second most suitable topic ensures that a wider choice of relevant fact-checked articles are considered for recommendation, while the combination of top topics help improve the accuracy of the recommendations to users with relevant and accurate information to counter misinformation (refer to the third category of recommendation in Section \ref{sec:approach}).


\textbf{Knowing the Unknown:} The \textbf{Unknown} topic label is linked with the topmost topics of articles, with the second most topic is the \textbf{Unknown}. There were a total of 21 fact-checked articles that were not being assigned a second topic (thus \textbf{Unknown}), as reported in Table \ref{tbl:topic_distribution_2}.
We explored those 21 \textbf{Unknown} articles manually. They discuss an assortment of themes such as throwing away God's idols into the river due to COVID-19, lockdown concerns, questioning the existence of COVID-19, training drills wrongly linked to COVID-19 havoc, mutation of human vs. man-made viruses, and black fungus correlated with COVID-19. 
Though these \textbf{Unknown} topics may not as popular as the other topics in the dataset, they are still vital, since they add diversity and coverage of fact-checking of a wider variety of misinformation. 

%% file: Proactive_approach.tex
\section{Repurposing Social Media Content}\label{sec:proactive_recommendations}

Our first approach to combating misinformation on social media is to utilize contents from the platform itself to recommend relevant non-misinformation posts in response to a misinformation post (refer to the top box in Figure \ref{fig:proposed}). This is particularly useful in situations where fact-checking articles might not (yet) be available.

To that end, we employ a combination of three 
Natural Language Processing (NLP) techniques. The first technique is Latent Dirichlet Allocation (LDA) topic modeling, which is used to extract general themes and topics from the posts (as elaborated previously in Section \ref{sec:social_topics}). We also use Named Entity Recognition\footnote{https://spacy.io/api/entityrecognizer} (NER) to identify specific entities mentioned in the posts (Section \ref{sec:ner}). Finally, we match the sentiments of the posts to ensure that the recommended posts provide a balanced perspective on the topic (Section \ref{sec:sentiments}). 




\subsection{Specific Topic Extraction}\label{sec:ner}

We evaluated a variety of Named Entity Recognition (NER) models, including the latest pretrained roberta-base NER model, which successfully identified named entities in 57\% of our tweets data. 
However, after careful examination, we found that the en\_core\_web\_sm model performed best, identifying entities in 88\% of our data with at least one entity.

Upon further investigation, we discovered a few issues with pre-existing models. Firstly, certain words specific to COVID-19, such as Pfizer, Shots, and Johnson \& Johnson, were incorrectly labeled as PERSON, GPE (Geopolitical entity). Secondly, there were still 12\% of entries that contained zero or null entities.

To address these issues, we fine-tuned the NER model on a subset of 100 manually labeled random instances from our dataset. We added a new entity type called VAC\_TYPE for vaccine names using a list of manually labeled VAC\_TYPE entities: [pfizer, astrazeneca, mrna, astrazenca, jnj, oxford, sputnik, modern, variants, \#pfizer, booster, \#astrazeneca, biontech, Covidshield]. 
Additionally, we also included spelling errors in the entities and did not remove hashtags from the dataset as they play a crucial role in identifying misinformation. We trained the augmented model for 30 epochs with a dropout rate of 0.3, using five-fold cross-validation.

We then compared the performance of the augmented model (fine-tuned) to the unaugmented model on the same subset of 100 examples, as shown in Table \ref{tbl:comparison}. We observed that the augmented model performed better in all metrics by a significant margin. Finally, we labeled all tweets in the dataset using the augmented model.
We obtained the list of VAC\_TYPE entities that are labeled by the augmented model, which included new entries beside the ones we had provided during training, viz. [phizer, myrna, zenca, novavax, johnsonandjohnson, johnson, mirna].

Through this process, we were able to improve the detection of entities and entity types in the dataset. In particular, for 112,994 
(98.5\%) of our records, at least one entity was identified (up from 88\% with the unaugmented one).
Besides identification of new entities across the dataset, 36,829 previously identified entities were reclassified by the augmented model: 35,480 (96.3\%) of these were of VAC\_TYPE and 1,349 (3.7\%) were of non-VAC\_TYPE. Table \ref{tbl:mislabeled} represents the top entities divided into  VAC\_TYPE and non-VAC\_TYPE entities. For example, the entity Pfizer has been incorrectly labeled (in the present context) as ORG by the unaugmented model but as VAC\_TYPE by the augmented model. Since most of the dataset contains vaccine names in their tweets, it clearly indicates why VAC\_TYPE entities detected by the model are in such a big number.

\begin{table}[!htbp]
\centering
\caption{Comparison of the Spacy NER model with and without fine-tuning the model on the manually labeled data.}
\label{tbl:comparison}
\resizebox{0.8\columnwidth}{!}{%
\begin{tabular}{|l|l|l|l|l|}
\hline
\textbf{Spacy model} & \textbf{Accuracy} & \textbf{Precision} & \textbf{Recall} & \textbf{F1 Score} \\ \hline
w/o training         & 0.27              & 0.25               & 0.26            & 0.25              \\ \hline
w/ training          & 0.89              & 0.90               & 0.88            & 0.87              \\ \hline
\end{tabular}%
}
\end{table}

While the augmented model generally performed better, there were a few instances where it failed to identify entities accurately. For example, in a few cases, geographic places were mislabeled as VAC\_TYPE by the augmented model but correctly labeled as GPE by the unaugmented model. Additionally, there were instances where both models failed to identify the entity accurately, for instance, the un/augmented models have incorrectly labeled Ohio as MONEY and VAC\_TYPE, respectively.
Despite these limitations, our approach resulted in a substantial improvement in the detection of named entities and entity types within the tweets.

\begin{table}[!htbp]
\footnotesize
\centering
\caption{Top entities mislabeled by the unaugmented NER model. The left and right sides represent vaccine names and non-vaccine names entities.
}
\label{tbl:mislabeled}
\resizebox{\columnwidth}{!}{%
\begin{tabular}{|ccc|ccc|}
\hline
\multicolumn{3}{|c|}{\textbf{VAC\_TYPE}}                                                                & \multicolumn{3}{c|}{\textbf{Others}}                                                               \\ \hline
\multicolumn{1}{|c|}{\textbf{Entity}}     & \multicolumn{1}{c|}{\textbf{Mislabeled}} & \textbf{Correct} & \multicolumn{1}{c|}{\textbf{Entity}} & \multicolumn{1}{c|}{\textbf{Mislabeled}} & \textbf{Correct} \\ \hline
\multicolumn{1}{|c|}{pfizer}              & \multicolumn{1}{c|}{ORG}                 & VAC\_TYPE        & \multicolumn{1}{c|}{millions}        & \multicolumn{1}{c|}{CARDINAL}            & MONEY            \\ \hline
\multicolumn{1}{|c|}{moderna}             & \multicolumn{1}{c|}{GPE}                 & VAC\_TYPE        & \multicolumn{1}{c|}{billions}        & \multicolumn{1}{c|}{CARDINAL}            & MONEY            \\ \hline
\multicolumn{1}{|c|}{astrazeneca}         & \multicolumn{1}{c|}{ORG}                 & VAC\_TYPE        & \multicolumn{1}{c|}{trump}           & \multicolumn{1}{c|}{ORG}                 & PERSON           \\ \hline
\multicolumn{1}{|c|}{johnson and johnson} & \multicolumn{1}{c|}{PERSON}              & VAC\_TYPE        & \multicolumn{1}{c|}{biden}           & \multicolumn{1}{c|}{ORG}                 & PERSON           \\ \hline
\multicolumn{1}{|c|}{novavax}             & \multicolumn{1}{c|}{ORG}                 & VAC\_TYPE        & \multicolumn{1}{c|}{lock down}       & \multicolumn{1}{c|}{NORP}                & EVENT            \\ \hline
\end{tabular}%
}
\end{table}

\subsection{Sentiments Extraction}
\label{sec:sentiments}

To counter confirmation bias, we use the VADER API \cite{hutto2014vader} to detect the sentiment of tweets as positive, negative, or neutral. To recommend related non-misleading tweets to counter misleading tweets, we employ two approaches for determining similarity: direct string matching and vector dimension matching. We found that vector dimension matching with GloVe embedding \cite{pennington2014glove} and cosine similarity worked best for our data. 

\subsection{Top-K non-misleading relevant tweets}
\label{sec:topktweets}

In a nutshell, we employed three criteria - general topic, entities extracted by our augmented NER model, and sentiment - to match misleading tweets with equivalent non-misleading tweets. The top-K (where K can be a user-defined parameter) most similar non-misleading tweets are then identified using the GloVe embedding and cosine similarity approach to be recommended as counters. 

\shakshi{
In the case of retrieving no similar non-misleading tweets for a given misleading tweet, we try to find the closest match of the misleading and non-misleading tweets using our three criteria, that is, topics, sentiments and entities. More precisely, we relax our criteria in this case by requiring a match of at least two entities, for instance. 
}

\subsection{Criterion for Evaluation of First Approach}\label{sec:evaluate_nonmisleading}
To evaluate our approaches, we employ the Mean Reciprocal Rank ($MRR$) and Mean Average Precision ($MAP$)  metrics, which are frequently employed in recommendations tasks \cite{wang2022veracity,liang2016modeling,haruna2017collaborative}, particularly in settings where ground truth is absent.

This is particularly pertinent since it is impractical to manually evaluate the recommendations' quality. 
Hence, we stick to the two widely used metrics that are suitable in our case.  Particularly, we adapted the definition of relevant items considered in both the mentioned metrics. 
To that end, we define Precision@k ($P@k$) as follows:

\begin{equation}
P@k = \frac{\#\; relevant\; recommendations\; to\; misleading\; tweet}{\#\; recommendations\; to\; misleading\; tweet} 
\end{equation}

Here, a recommendation is considered relevant if, after similarity matching of misleading and non-misleading tweet pairs, the non-misleading tweets satisfy all three criteria: matching topics, entities, and sentiments with the misleading tweet. 
The total number of recommendations involves non-misleading tweets recommendations utilizing similarity measure from the whole corpus (that is, without any criteria).
The rest of the $MAP$ calculation is the same as per the predefined formula \cite{liang2016modeling}.
\begin{equation}
MAP@k = \frac{1}{N}
    \sum_{i=1}^{N}\frac{1}{K} \sum_{k=1}^{K} P@k \cdot rel(k)
\end{equation}
where $rel(k)$ represents whether the recommendation is relevant or not. \shakshi{Thus, $rel(k)$ returns 1 if the recommendation of non-misleading tweet is relevant, i.e., if non-misleading tweets satisfy all three criteria mentioned above. Otherwise, $rel(k)$ returns 0. }
The $MRR$ metric remains the same as in \cite{xu2016tag}.
\begin{equation}
MRR@k = \frac{1}{R}
    \sum_{r=1}^{R}\frac{1}{k_{r}}
\end{equation}
where $k_{r}$ is the rank of the first relevant recommendations. $R$ is the number of misleading tweets that we seek recommendations for.

Table \ref{tbl:evaluation_metric}, row 1 represents the evaluation metrics $MRR@k$ and $MAP@k$ performances on this social media non-misleading (true) posts recommendations approach. It can be observed that at first, with an increasing value of k, performance starts to increase and then decreases.  $MRR@10$ and $MAP@10$ achieve better performance meaning that ten non-misleading posts recommendations are good enough to be presented as a counter for the misleading post. 
Furthermore, giving users too many recommendations (though ``too many" is a subjective word here) may cause them to lose interest in reading further.

\begin{table*}[!htbp]
\centering
\caption{Evaluation of the complementary approaches. AMIR\_SM and AMIR\_FC represent social media non-misleading posts recommendations (first approach) and fact-checked articles recommendations (second approach), respectively.}
\label{tbl:evaluation_metric}
\begin{tabular}{l|l|l|l|l|l|l|l|l|l|l|}
\cline{2-11}
\textbf{}                               & \textbf{MRR@3} & \textbf{MRR@5} & \textbf{MRR@10} & \textbf{MRR@15} & \textbf{MRR@20} & \textbf{MAP@3} & \textbf{MAP@5} & \textbf{MAP@10} & \textbf{MAP@15} & \textbf{MAP@20} \\ \hline
\multicolumn{1}{|l|}{\textbf{AMIR\_SM}} &     0.541           &           0.616     &       \textbf{0.689}           &   0.615            &      0.612           & 0.528          & 0.567          & \textbf{0.746}  & 0.703           & 0.677           \\ \hline
\multicolumn{1}{|l|}{\textbf{AMIR\_FC}} &    0.513            &       0.553       &   0.586              &      \textbf{0.663}            &     0.617           & 0.586          & 0.683          & 0.729           & \textbf{0.757}  & 0.738           \\ \hline
\end{tabular}%
\end{table*}




%% file: Factcheck_approach.tex

\section{Leveraging Fact-Checked Articles}\label{sec:approach}


\subsection{Association of topics (topic-topic mapping)}

First, we map the topics identified across both corpora to be able to identify a shortlist of fact-checked article(s) dataset which may share relevance to a given misleading tweet.

\textbf{Distance Metric Approach:} \shakshi{We use the Pyldavis API\footnote{https://pyldavis.readthedocs.io/en/latest/} which considers the top two principal components of topic distances using the Jensen-Shannon divergence \cite{menendez1997jensen} as a way to quantify the difference between two distributions. This metric has been widely used as a powerful measure for comparing the distributions in misinformation recommendation domains \cite{vrijenhoek2022radio,aires2020information,vrijenhoek2023radio}.


The approach works as following. For each topic in the misleading Twitter corpus, we find its coordinates among the topics in the FaCov corpus. To determine the closest match between one misleading topic and all the FaCov topics, we calculate the Euclidean distance between them. The closest match (that is, the smallest distance) between a misleading topic and a FaCov topic is then mapped together for topic-topic mapping. 
This analysis is performed for all misleading Twitter corpus' topics, resulting in 15 topic-topic mappings (shown in Table \ref{tbl:facov_subtopics_mapping_misleading_topics}).
}

We also tried a few variants to directly match strings of the keywords extracted from LDA in both corpora, which are as follows:

\begin{enumerate}
\item  \textbf{Naive approach:} We first removed stop words, lowercasing, and stemming of all the keywords extracted from LDA in both datasets. Then, examining the keywords extracted from the LDA topic modeling analysis for each topic, if there were any exact matching words between two topics, we considered their frequency of occurrence and sort them in a descending order. 
We repeated this analysis with all the topics' keywords in the second dataset by fixing the topic of one dataset. The final decision for mapping is based on the number of matched words and their frequency of occurrence, resulting in 15 topic-topic mappings which turned out to be identical to the mappings obtained through the measurement of distance of distributions.

\item \textbf{Term Frequency–Inverse Document Frequency (TF-IDF) weighted approach:} To find the mapping of the two topics from the datasets, the TF-IDF\footnote{\url{https://scikit-learn.org/stable/modules/generated/sklearn.feature\_extraction.text.TfidfVectorizer.html}} of each keyword in the topics is identified and used to find the cosine similarity of the topics. 
The problem with this approach was that it returned quite low similarity scores when the matching number of keywords is few, which was often the case, rendering it inapplicable. 

\item \textbf{Without TF-IDF weighted approach:} 
Cosine similarity scores without weighting using the spacy API\footnote{https://spacy.io/api/doc} yielded marginally better results, but failed to accurately reflect similarity of COVID-19 related terms, such as a 0 score between covid and mask.
\end{enumerate}

We manually validated all the topic-topic mapping to ensure the correctness of the results, and for the given data sets, we observed an accuracy of 100\% for the mappings.
We also evaluated the quality of the topic-topic mapping using the \textbf{Rank-k} approach. 
Particularly, for the naive approach which had identical result as the distance metric approach, we investigated the rank of each exact word (sorted in descending order if the matched words are more than one) that is mapped in all the topic-topic mappings. The value of k was determined by averaging all the ranks retrieved from repeating the approach for all the mappings. 
We observed that the exact words that are used to map the two topics appeared in the top 5 occurrences, so the rank was 5. 
Since the rank is quite well, we conclude that the naive approach is able to map the topics fast and hence, does not degrade the quality of the approach.

Given the disparity in the number of topics in each corpus, multiple topics of the misleading corpus mapped to one topic in the FaCov corpus, as shown in Table \ref{tbl:facov_subtopics_mapping_misleading_topics}. \textcolor{blue}{Blue} color corresponds to the topics of the Twitter dataset being mapped to the topics (black color) of the FaCov dataset.
For instance, \textbf{Vaccine's effects} has been mapped to multiple topics such as \textbf{Vaccine Efficacy, Shots, Choices}, etc.

Three topics of the Twitter dataset have not been mapped to any of the topics in the FaCov dataset, namely, \textbf{Operation Warp Speed, Real Side-effects, and Vaccine Approval}.
We further dig deep into each of the three topics, as can be seen in Tables \ref{tbl:ows} and \ref{tbl:not_mapped_side_effects}.

In the fact-checked articles of these three topics we discovered that the \textbf{Operation Warp Speed, and Real Side-effects} are mentioned very infrequently. As such, these are not main topics of the fact-checked articles, hence, these topics are not detected in the FaCov dataset even though these words are sub-parts that are being used to explain the main topic of the articles.
We then looked into these articles' topmost and second most topics.

In the 14 Operation Warp Speed articles (Table \ref{tbl:ows}), the most discussed topics are vaccine effects and trump-related posts. Similarly, in the  12 articles of the Real side-effects keyword, the most discussed topic is the vaccine effects.
However, in Vaccine Availability, we found a total of 185 fact-checked articles, including false positive cases (not enumerated due to space constraints).  

\begin{table}[!htbp]
\centering
\caption{Not mapped topic: Operation Warp Speed. Topic 1 and Topic 2 includes two prominent topics that mentions operation warp speed briefly. The rows indicate the number of social media posts where operation warp speed topic has been mentioned briefly}
\label{tbl:ows}
\resizebox{0.8\columnwidth}{!}{%
\begin{tabular}{|l|l|l|}
\hline
\textbf{No.} & \textbf{Topic 1}                  & \textbf{Topic 2}                                                         \\ \hline
1.           & Vaccine Effects                   & Masks                                                                    \\ \hline
2.           & Masks                             & \begin{tabular}[c]{@{}l@{}}Trump posts on \\ health Workers\end{tabular} \\ \hline
3.           & Trump posts on health Workers     & Vaccine Effects                                                          \\ \hline
4.           & Trump posts on health Workers     & Vaccine Effects                                                          \\ \hline
5.           & Vaccine Effects                   & \begin{tabular}[c]{@{}l@{}}Trump posts on \\ health Workers\end{tabular} \\ \hline
6.           & Trump posts on health Workers     & Spread of COVID-19                                                       \\ \hline
7.           & Trump posts on health Workers     & Vaccine Effects                                                          \\ \hline
8.           & Vaccine Effects                   & \begin{tabular}[c]{@{}l@{}}Trump posts on \\ health Workers\end{tabular} \\ \hline
9.           & Trump posts on health Workers     & Masks                                                                    \\ \hline
10.          & Vaccine Effects                   & Masks                                                                    \\ \hline
11.          & Vaccine Effects                   & Masks                                                                    \\ \hline
12.          & Vaccine Effects                   & Masks                                                                    \\ \hline
13.          & Masks                             & Vaccine Effects                                                          \\ \hline
14.          & Coronavirus death/cases in states & Vaccine Effects                                                          \\ \hline
\end{tabular}
}
\end{table}

\begin{table}[!htbp]
\centering
\caption{Not mapped topic: Real Side-effects. Topic 1 and Topic 2 includes two prominent topics that mentions operation warp speed briefly. The rows indicate the number of social media posts where operation warp speed topic has been mentioned briefly}
\label{tbl:not_mapped_side_effects}
\resizebox{0.8\columnwidth}{!}{%
\begin{tabular}{|l|l|l|}
\hline
\textbf{No.} & \textbf{Topic 1}         & \textbf{Topic 2}                                \\ \hline
1.  & Vaccine Effects & Masks                                  \\ \hline
2.  & Vaccine Effects & Masks                                  \\ \hline
3.  & Vaccine Effects & -                                      \\ \hline
4.  & Vaccine Effects & Misleading posts, pictures, and videos \\ \hline
5.  & Vaccine Effects & Trump posts on health Workers          \\ \hline
6.  & Vaccine Effects & -                                      \\ \hline
7.  & Vaccine Effects & Coronavirus death/cases in states      \\ \hline
8.  & Vaccine Effects & Misleading posts, pictures, and videos \\ \hline
9.  & Vaccine Effects & Trump posts on health Workers          \\ \hline
10. & Vaccine Effects & Masks                                  \\ \hline
11. & Vaccine Effects & Masks                                  \\ \hline
12. & Vaccine Effects & Trump posts on health Workers          \\ \hline
\end{tabular}
}
\end{table}

\subsection{Filtering the fact-checked articles}

After topic-topic mapping, for a given misleading tweet from a specific topic, its mapped with the topic of the FaCov dataset utilizing Table \ref{tbl:facov_subtopics_mapping_misleading_topics} to filter out irrelevant fact-checked articles and shortlist only those articles that are specific to the misleading tweet's topic.


\subsection{Semantic Matching}

The filtered fact-checked articles for a given misinformation tweet are then used to carry out semantic matching using a pre-trained Sentence Textual Similarity transformer model \cite{reimers2019sentence}. Specifically, each [tweet, article] pair is fed to the pre-trained model which outputs their cosine similarity scores. These scores are sorted in descending order to analyze whether the most relevant fact-checked article(s) from the FaCov corpus for a given misleading tweet can be recommended or not.

We used two pre-trained similarity transformer models that are popular and trained on large and multiple datasets, including social media datasets, namely,  \textit{all-MiniLM-L6-v2} and \textit{paraphrase-MiniLM-L6-v2} from the API\footnote{https://www.sbert.net/index.html}. In our case, the \textit{all-MiniLM-L6-v2} model is the best performing model as per manual validation.
One big advantage of these text similarity models is that they also take care of the contextual (or synonymous) words while comparing the pair. 
Next, we investigated the following questions while performing the similarity task.

\begin{enumerate}
    \item \textbf{Title or content?} We performed the text similarity on both types of pairs, that is, [tweet, title column of the article] and [tweet, content column of the article]. The \textit{title} and \textit{content} columns are the title of the fact-checked article and the whole content of the fact-checked article, respectively. We found that the \textit{title} column performs better than the \textit{content} column. This could be because tweets are short in length, so comparing the tweet and the title makes more sense as compared to the tweet and content, which has many more words and that could hinder the matching process. We performed our analysis with all the different combinations of topics. From the Twitter corpus, we picked a misleading tweet from a specific topic.

\item \textbf{How to pick a misleading tweet?} We considered the engagement analysis, which includes the replies count, retweets count, and likes count of a tweet. The premise is whether if the engagement analysis of a misleading tweet is high, it is more likely to find relevant fact-checked articles in the FaCov corpus. 
We checked by picking the highest and lowest engagements of the same topic's misleading tweets and found no such correlation between high engagement and the presence of relevant fact-checked articles. 
Thus, the misleading tweet of a specific topic has been chosen randomly.

\end{enumerate}

\subsection{The Three Customized Recommendations}

Next, after mapping the topics of the FaCov corpus, we perform the text similarity and find the cosine similarity scores.
For all the various topic combinations, we observe that the similarity scores do not exceed 69\% in the corpora.
By analyzing these combinations, we surmise the following categories of recommendations:
\begin{enumerate}
    \item \textbf{Specific recommendations: }If the similarity scores are equal to or greater than 62\%, 
    then the specific recommendations of the article(s) are provided to a misleading tweet. The threshold of 62\% has been decided by manual validation. 
    \shakshi{The process involved validating with different threshold values manually and analyzing the results to see if the recommendations provided were specific and relevant to the misleading tweet. Specifically, we noted the range of highest (69\%) and lowest similarity scores (20\%) retrieved after the semantic matching approach (as discussed in Section \ref{sec:approach} C). Next, we randomly picked 100 subsets of [misleading tweet, fact-check article] pairs from our data covering all the topics and determined if the similarity score is equal to or above 62\%, in which case it was deemed that the fact-check article is discussing the same topics, entities, and sentiments. Otherwise, if the score was less than 62\%, we noted that the entities or the sentiments started to deviate. Hence, they would not fit the first type of recommendation.}

    When we refer to specific recommendations, we mean that the recommended articles should have the exact same topic as the given misleading tweet. The user will be able to access articles that are closely relevant to the information stated in the misleading tweet in this way, giving them access to more reliable and correct information. 
    For instance, the top recommended article for the misleading tweet \textit{``they all used aborted fetus either in development or testing. Especially, in the vaccine astrazeneca johnson and johnson have fetal cell lines and is being said the mrna is something never tested on humans''} from the topic \textbf{Vaccine Efficacy} is \textit{``Johnson \& Johnson's COVID-19 vaccine does not contain aborted fetal cells''} with 65\% similarity score.
    \item \textbf{Almost near recommendations: }If the similarity score is less than 62\%, then we recommend such articles that have an almost near topic discussed in the given misleading tweet. \shakshi{This means that if any of the two criteria, that is, sentiments or entities retrieved from Section \ref{sec:approach} C, differs between the [misleading tweet, fact-check article] pairs (which is less than 62\%), then this second type of recommendation is considered.}
    For instance, the top recommended article for the misleading tweet \textit{``i am proud of you for refusing it. Astrazeneca vaccine is dubious too many bad cases than goods in results of taking that, so whether it works or not it won't be any of my portion''}  from the topic \textbf{Choices} is \textit{``Experts say the Oxford AstraZeneca COVID-19 vaccine is safe and that its benefits far outweigh possible risks''} with 59\% similarity score.
    \item \textbf{General/Broad recommendations: }\shakshi{In a case where there is no mapping of the tweet's topic with the fact-checked topic, that is, zero articles have been retrieved from the fact-checked corpus, the second best prominent topic (discussed in Section III-C) can be used to find the similarity score which can further be employed for broad recommendations. 
    Using the second most suitable topic ensures that all relevant fact-checked articles are considered for recommendation. This can help to improve the coverage of the recommendations and ensure that users are provided with relevant and accurate information to counter misinformation in a best effort manner. 
    For instance, the 
    \textbf{School e-learning} topic of FaCov has no mapping with the \textbf{Operation Warp Speed} of misleading corpus. In this case, the second prominent topic in the FaCov corpus, which is, \textbf{Trump posts on health workers} (as per Figure \ref{fig:graph_example})  can be associated with the \textbf{Operation Warp Speed} topic in the misleading corpus.
    For instance, the top recommended article for the misleading tweet \textit{``maybe he kept some so he can sell them to the black market remember he is in debt or to punish BioNTech pfizer for letting the world know that operation warp speed money was not involved in their vaccine development. This mean guy is capable of heinous crimes''}  from the topic \textbf{Operation Warp Speed} is \textit{``No, Trump didn't tweet his blood is a vaccine''} with a 37\% similarity score.}
\end{enumerate}

\subsection{Evaluation Criterion of Second Approach}\label{sec:evaluate_fc}

To evaluate our second approach, which is recommending fact-checked articles to a misleading post, we employed the $MRR@k$ and $MAP@k$ metrics as discussed in Section \ref{sec:evaluate_nonmisleading}. Other widely used metrics such as $NDCG@k$ require the availability of ground truth availability which is not the case for our problem.
Table \ref{tbl:evaluation_metric}, row 2 shows the performance of the $MRR@k$ and $MAP@k$ metrics for different values of k. It can be seen from the table that 15 fact-checked articles' recommendations for a misleading post provide good performance. Further, it is to be noted the higher value of $k$ does not necessarily mean better performance as we may not have the relevant recommendations to provide to the social media user. Additionally, giving the user too many recommendations can make them less effective due to limited attention. Thus, the focus here is on the quality rather than the quantity of the recommendations.

%% file: conclusions.tex
\section{Conclusion}\label{sec:conclusion}

In this work, we demonstrate the feasibility of automated misinformation rebuttal systems to battle misinformation on social media, amortizing user generated content from the same platform in conjunction with articles from websites reputed and specializing in fact-checking popular misinformation.

\shakshi{The whole pipeline involves several complex tasks, e.g., discriminating social media content carrying misinformation from those that do not, understanding the broad and precise topic of the content, correlating with the content of fact-checked articles when applicable, in order to determine relevance of recommendations. Since the thrust of this work was on recommendations, we disentangled it from the initial task of classification, and instead leveraged existing curated data sets, and confined our study to Covid-19 related topics within only Twitter posts in English. 

It is thus of interest to generalize our work across several dimensions - coverage of broad topics beyond Covid-19, considering content across a wider range of open social media platforms such as Reddit and addressing non-English content. Furthermore, our current approach leads to recommended content which are individually from only a single source. Using large language models to synthesize recommendations amalgamating content from multiple sources, and being able to do so even in non-English languages would immensely enhance the utility of such a rebuttal system.   }

While we have used manual inspections and human interpretation of the intermediate results, since all the actual steps in the pipeline are algorithmic in nature, our approach can eliminate human-in-the-loop and in that sense, in principle, it can function at scale, and in near real time. Nevertheless, a real live deployment dealing with a plethora of sources would need to optimize the various algorithms for computational efficiency in addition to any work to improve the accuracy of the results. For example, right now, in Section \ref{sec:approach} we use what can be deemed as an exhaustive search for establishing topic-topic maps for data from the two different data sources, which is undesirable, more so when we would need to accommodate a wider set of topics and data sources. \shakshi{As such, while the system operates in a fully automated manner in making recommendations, there is still need to enhance the degree of automation in the back-end data-analytics-pipeline.} 

\shakshi{Finally, the strategy outlined in this contribution focuses primarily on the specific topic of countering COVID-19 misinformation on Twitter. In that context, a bottleneck was the lack of ground truth regarding the availability and quality of recommendations to be found within the specific collection of user generated corpus. This prevented us from using a wider set of evaluation metrics to evaluate the quality of recommendations. }

Hence, we focus on the two widely used metrics by adapting the definition to fit our scenarios to evaluate our complementary approaches. This, along with longitudinal user case studies employing A/B tests on the efficacy of even the premise itself, namely timely and precise rebuttals might thwart misinformation (and to which extent) are open ended issues that need further investigation.